\pdfoutput=1

\documentclass[letterpaper, 10pt, conference]{ieeeconf}   %

\IEEEoverridecommandlockouts                              %

\let\labelindent\relax                                    %

\usepackage{graphicx}     %
\usepackage{amsmath}      %
\usepackage{amssymb}      %
\usepackage{amsfonts}     %
\usepackage{mathtools}    %
\usepackage{enumitem}     %
\usepackage{siunitx}      %
\usepackage{tensor}       %
\usepackage{url}          %
\usepackage{xcolor}       %
\usepackage{xspace}       %
\usepackage[T1]{fontenc}  %
\usepackage{hyperref}     %
\usepackage[all]{hypcap}  %

\hypersetup{hidelinks, bookmarksdepth=3, bookmarksopen, bookmarksnumbered, colorlinks=false, linkcolor={red!50!black}, citecolor={blue!50!black}, urlcolor={blue!80!black}}

\makeatletter
\def\subsubsection{\@startsection{subsubsection}{3}{\parindent}{0ex plus 0.1ex minus 0.1ex}{0ex}{\normalfont\normalsize\bfseries}}
\makeatother

\makeatletter
\newcommand*\bigcdot{\mathpalette\bigcdot@{.5}}
\newcommand*\bigcdot@[2]{\mathbin{\vcenter{\hbox{\scalebox{#2}{$\m@th#1\bullet$}}}}}
\makeatother

\graphicspath{{images/}}
\DeclareGraphicsExtensions{.pdf,.png,.jpg,.jpeg}

\DeclareMathOperator{\sign}{sign}
\DeclareMathOperator{\asin}{asin}
\DeclareMathOperator{\acos}{acos}
\DeclareMathOperator{\wrap}{wrap}
\DeclareMathOperator{\slerp}{slerp}
\DeclareMathOperator{\atantwo}{atan2}

\newcommand{\vect}[1]{\mathbf{#1}} %
\newcommand{\vecs}[2]{\tensor*{\vect{#1}}{_{#2}}} %
\newcommand{\vecb}[2]{\tensor*[^{#1}]{\vect{#2}}{}} %
\newcommand{\vecbs}[3]{\tensor*[^{#1}]{\vect{#2}}{_{#3}}} %
\newcommand{\compb}[2]{\tensor*[^{#1}]{#2}{}} %
\newcommand{\compbs}[3]{\tensor*[^{#1}]{#2}{_{#3}}} %
\newcommand{\rots}[2]{\tensor*{#1}{_{#2}}} %
\newcommand{\rotb}[3]{\tensor*[^{#1}_{#2}]{#3}{}} %
\newcommand{\rotbs}[4]{\tensor*[^{#1}_{#2}]{#3}{_{#4}}} %
\newcommand{\quatb}[3]{\tensor*[^{#1}_{#2}]{#3}{}} %
\newcommand{\yawof}[1]{\Psi\bigl(#1\bigr)} %
\newcommand{\fr}[1]{\{#1\}\xspace} %
\newcommand{\inv}{^{-1}} %
\newcommand{\trans}{^T\!} %
\newcommand{\I}{\mathbb{I}} %
\renewcommand{\P}{\mathbb{P}} %
\newcommand{\Q}{\mathbb{Q}} %
\newcommand{\R}{\mathbb{R}} %
\newcommand{\T}{\mathbb{T}} %
\newcommand{\half}{\tfrac{1}{2}} %
\newcommand{\hpi}{\tfrac{\pi}{2}} %
\newcommand{\defeq}{\equiv} %
\newcommand{\degreem}{^{\circ}} %
\newcommand{\dotp}{\bigcdot} %

\newcommand{\pxt}{\tilde{p}_x}
\newcommand{\pyt}{\tilde{p}_y}
\newcommand{\pzt}{\tilde{p}_z}
\newcommand{\pxd}{\dot{p}_x}
\newcommand{\pyd}{\dot{p}_y}
\newcommand{\pzd}{\dot{p}_z}
\newcommand{\pxtd}{\dot{\tilde{p}}_x}
\newcommand{\pytd}{\dot{\tilde{p}}_y}
\newcommand{\pztd}{\dot{\tilde{p}}_z}
\newcommand{\gammat}{{\tilde{\gamma}}}
\newcommand{\gampsi}{{\gamma + \psi}}

\newcommand{\seclabel}[1]{\label{sec:#1}}

\newcommand{\figlabel}[1]{\label{fig:#1}}

\newcommand{\eqnlabel}[1]{\label{eqn:#1}}

\newcommand{\secref}[1]{Section~\ref{sec:#1}\xspace}

\newcommand{\figref}[1]{Fig.~\ref{fig:#1}\xspace}

\newcommand{\eqnref}[1]{(\ref{eqn:#1})\xspace}

\newcommand{\secrefs}[2]{Section~\ref{sec:#1}--\ref{sec:#2}\xspace}
\newcommand{\eqnrefs}[2]{(\ref{eqn:#1}--\ref{eqn:#2})\xspace}

\newcommand{\cpp}{C\texttt{\nolinebreak\hspace{-.05em}+\nolinebreak\hspace{-.05em}+}\xspace}
\newcommand{\degree}{$\degreem$\xspace}

\title{\LARGE \bf Tilt Rotations and the Tilt Phase Space}

\author{Philipp Allgeuer and Sven Behnke%
\thanks{All authors are with the Autonomous Intelligent Systems (AIS) Group, Computer Science Institute VI,
        University of Bonn, Germany. Email: {\tt\small pallgeuer@ais.uni-bonn.de}. This work was partially
        funded by grant BE 2556/13 of the German Research Foundation (DFG).}}

\usepackage{eso-pic}

\AtBeginDocument{\AddToShipoutPictureFG*{\AtTextUpperLeft{\put(0,\LenToUnit{9pt}){\parbox{\textwidth}{\centering\bfseries
International Conference on Humanoid Robots (Humanoids), Beijing, China, 2018
}}}}}
\begin{document}
\begin{allowdisplaybreaks}

\bstctlcite{IEEEexample:BSTcontrol}

\maketitle
\thispagestyle{empty}
\pagestyle{empty}

\begin{abstract}
In this paper, the intuitive idea of tilt is formalised into the rigorous 
concept of tilt rotations. This is motivated by the high relevance that pure 
tilt rotations have in the analysis of balancing bodies in 3D, and their 
applicability to the analysis of certain types of contacts. The notion of a 
`tilt rotation' is first precisely defined, before multiple parameterisations 
thereof are presented for mathematical analysis. It is demonstrated how such 
rotations can be represented in the so-called tilt phase space, which as a 
vector space allows for a meaningful definition of commutative addition. The 
properties of both tilt rotations and the tilt phase space are also extensively 
explored, including in the areas of spherical linear interpolation, rotational 
velocities, rotation composition and rotation decomposition.
\end{abstract}

\section{Introduction}
\seclabel{introduction}

Tilt rotations first arose in the context of the fused angles and tilt angles 
rotation representations, as a way of partitioning 3D rotations into independent 
yaw and tilt components \cite{Allgeuer2015a}. They are highly relevant for 
situations like the analysis of balancing bodies in 3D---e.g.\ bipedal robots 
\cite{PhaseFeed} \cite{Allgeuer2016a}, as the yaw component does not contribute 
to the body-local state of balance---but can also be used for other purposes. 
For example, tilt rotations can be used to represent the relative orientations 
of contacting bodies, or the relative rotations of two planar surfaces, such as 
for instance the foot of a robot and the ground. This paper seeks to 
comprehensively investigate tilt rotations, the parameterisations and properties 
they have, and their possible algorithmic applications. As part of this, the 
novel tilt phase space representation is introduced, which amongst other things 
allows for tilt rotations to be commutatively combined in a way akin to 
addition. Complete definitions of tilt rotations and the tilt phase space can be 
found in \secrefs{defn_tilt_rotations}{defn_tilt_phase}, but as an illustrative 
guide, \figref{tiltrot_teaser} shows three different tilt rotations applied to a 
robot, along with the corresponding axes of rotation. For tilt rotations, by 
definition the axis of rotation \emph{must} lie in the horizontal $\vect{xy}$ 
plane, as shown, and plotting the magnitude of the rotation at the polar angle 
corresponding to the axis of rotation gives the three 2D tilt phase space points 
shown on the right hand side in \figref{tiltrot_teaser}. When combined with a 
yaw component, e.g.\ to form the 3D tilt phase space, tilt rotations can be used 
to quantify and analyse any general 3D rotation.

In addition to the formalisation of the idea of referenced rotations (see 
\secref{referenced_rotations}), the main contributions of this paper lie in the 
systematisation of the concept of tilt rotations, the introduction of the novel 
tilt phase space, and the investigation of the many properties and results of 
both. Open source software libraries in \cpp and Matlab\footnote{\scriptsize 
\hspace{1pt}\cpp$\mspace{-6mu}/$Matlab: 
\url{https://github.com/AIS-Bonn/rot_conv_lib}\\ 
\url{https://github.com/AIS-Bonn/matlab_octave_rotations_lib}} have also been 
released to explicitly support tilt rotations, and computations and conversions 
involving the tilt phase space.

\begin{figure}[!t]
\parbox{\linewidth}{\centering\includegraphics[width=1.0\linewidth]{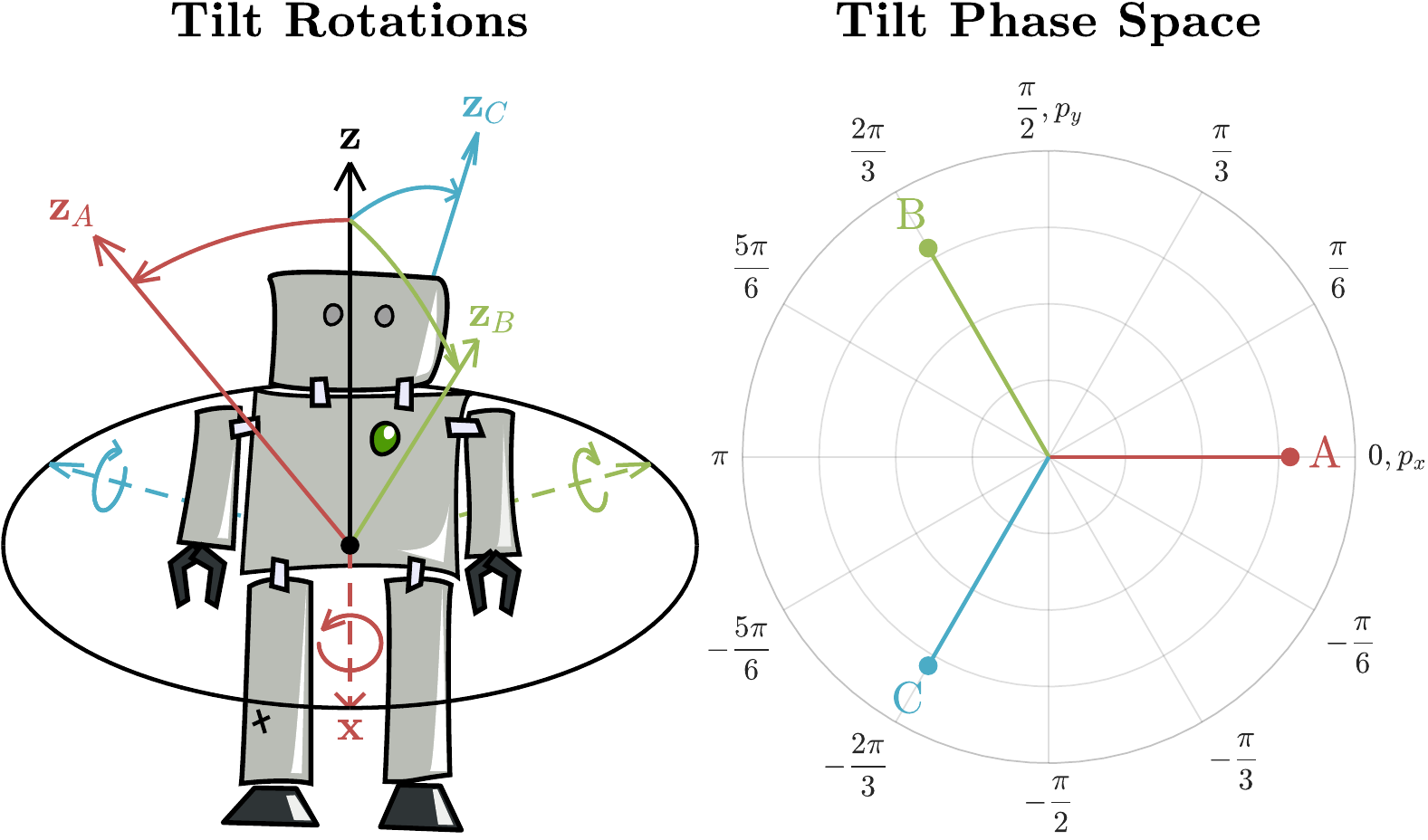}}
\caption{Diagram showing three different example tilt rotations A, B and C 
(left), and the corresponding points in the tilt phase space (right). More 
details on the definitions of both concepts follow in 
\secrefs{defn_tilt_rotations}{defn_tilt_phase}.}
\figlabel{tiltrot_teaser}
\vspace{-2ex}
\end{figure}
\section{Related Work}
\seclabel{related_work}

Tilt rotations have been used in previous works, for example for the modelling 
of heading-independent balance states while walking \cite{Allgeuer2016a}, and 
for the modelling of foot orientations and ground contacts \cite{Farazi2016}. 
The tilt phase space has also been used as the fundamental basis for an entire 
bipedal walking feedback controller \cite{PhaseFeed}, which was possible only 
due to its unique set of properties. Tilt rotations were first formulated as 
part of the definition of fused angles and tilt angles, but were not 
significantly further investigated in their own right. These two 
representations, in addition to the tilt phase space representation presented 
here, were developed to provide a robust and geometrically intuitive way of 
quantifying the components of rotation of a body in each of the three major 
planes, namely the $\vect{xy}$, $\vect{yz}$ and $\vect{xz}$ planes 
\cite{Allgeuer2015a}. This is something that quaternions and Euler angles do not 
do, as discussed in extensive detail in \cite{Allgeuer2018a}. As a result, any 
attempt to define the concept of a `tilt rotation' for example as the 
combination of Euler pitch and roll, would be mathematically unhelpful, and not 
correspond to human intuition of `tilt'. A thorough review of all kinds of 
rotation representations, such as for example quaternions, rotation matrices, 
axis-angle pairs \cite{Palais2009} and vectorial parameterisations 
\cite{Bauchau2003} \cite{Trainelli2004}, can be found in \cite{Allgeuer2015a}. 
The representation most closely related to the tilt phase space is the rotation 
vector \cite{Argyris1982}. Two of the tilt phase parameters can be expressed as 
the rotation vector parameters of the \emph{tilt rotation component} (see 
\secref{defn_fused_yaw}) of a rotation. There is no correspondence to the 
rotation vector parameters of the full rotation however.

\section{Preliminaries}
\seclabel{preliminaries}

This section introduces the notation and basic identities that are used 
throughout the remainder of this paper.

\subsection{Notation and Conventions}
\seclabel{notation}

As is standard for the analysis of fused angles and tilt angles, we use the 
convention that the z-axis points `up', where this is generally defined to be in 
the opposite direction to gravity, or along a particular surface normal, as 
required. The global fixed frame is taken to be \fr{G}, and the body-fixed frame 
of interest is taken to be \fr{B}. The rotation $\rotb{G}{B}{R}$ refers to the 
rotation from \fr{G} to \fr{B}, namely
\begin{equation}
\setlength{\arraycolsep}{2pt}
\rotb{G}{B}{R} =
\begin{bmatrix}
\vecbs{G}{x}{B} & \vecbs{G}{y}{B} & \vecbs{G}{z}{B}
\end{bmatrix} =
\begin{bmatrix}
\vecbs{B}{x}{G} & \vecbs{B}{y}{G} & \vecbs{B}{z}{G}
\end{bmatrix}\trans\!,
\eqnlabel{rotmatrowcolvec}
\end{equation}
where for example $\vecbs{G}{z}{B}$ is the unit vector corresponding to the 
positive z-axis of \fr{B}, expressed in the coordinates of \fr{G}. We further 
follow the convention that, for example,
\begin{equation}
\vecbs{G}{z}{B} = \bigl( \compbs{G}{z}{Bx}, \compbs{G}{z}{By}, \compbs{G}{z}{Bz} \bigr). \eqnlabel{axiscomponents}
\end{equation}
If the superscript basis frame is omitted, e.g.\ like in `$\vecs{z}{B}$', by 
default it is the global fixed frame \fr{G}.

\begin{figure}[!t]
\parbox{\linewidth}{\centering\includegraphics[width=1.0\linewidth]{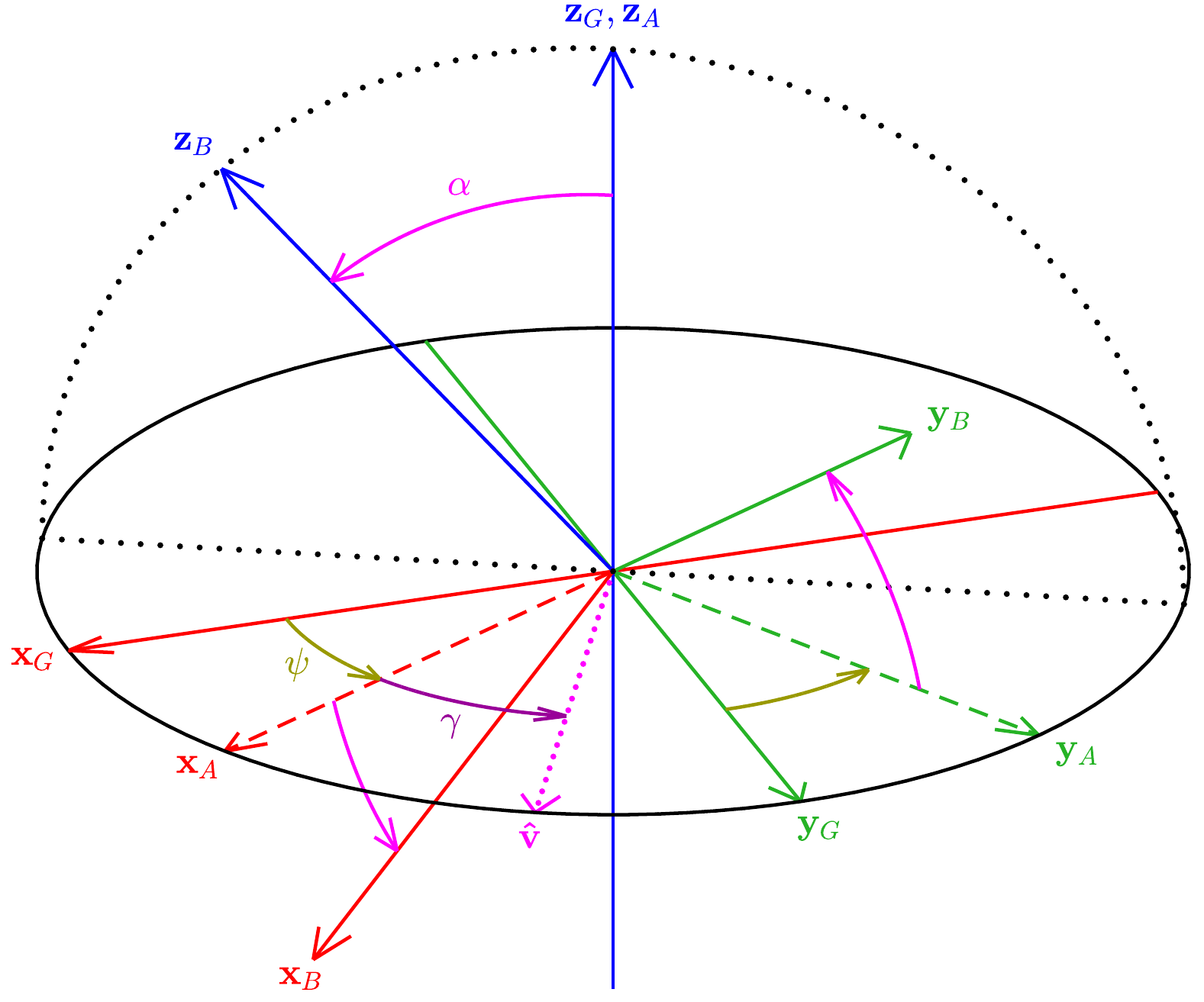}}
\caption{Diagram of the tilt angles parameters $(\psi, \gamma, \alpha)$. A 
z-rotation by $\psi$ from \fr{G} to \fr{A} is followed by a rotation by $\alpha$ 
about $\vect{\hat{v}}$ from \fr{A} to \fr{B}.}
\figlabel{tilt_parameters}
\vspace{-2ex}
\end{figure}
\subsection{Referenced Rotations}
\seclabel{referenced_rotations}

Given frames \fr{A} and \fr{B} relative to a global frame \fr{G}, the global 
rotation that maps \fr{A} onto \fr{B} is given by
\begin{equation}
\rotb{GA}{B}{R} = \rotb{G}{B}{R} \, \rotb{A}{G}{R}, \eqnlabel{refrotdefn2}
\end{equation}
where $\rotb{GA}{B}{R}$ is new notation for the rotation from frame \fr{A} to 
\fr{B} \emph{referenced by} \fr{G}. Alternative mathematical formulations for 
this so-called \emph{referenced rotation} include
\begin{equation}
\rotb{GA}{B}{R} = \rotb{G}{A}{R} \, \rotb{A}{B}{R} \, \rotb{G}{A}{R}\trans = \rotb{G}{B}{R} \, \rotb{A}{B}{R} \, \rotb{G}{B}{R}\trans. \eqnlabel{refrotdefn3}
\end{equation}
It should be noted that referenced rotations are just a direct generalisation of 
standard rotations, as from \eqnref{refrotdefn2} we have
\begin{align}
\rotb{G}{B}{R} &\equiv \rotb{GG}{B}{R}, & \rotb{A}{G}{R} &\equiv \rotb{GA}{G}{R}, & \rotb{GA}{A}{R} &= \I. \eqnlabel{refrottrivial}
\end{align}
Basic identities involving referenced rotations include
\begin{align}
\rotb{GA}{B}{R} \, \rotb{G}{A}{R} &= \rotb{G}{B}{R}, & \rotb{G}{A}{R}\trans \, \rotb{GA}{B}{R} \, \rotb{G}{A}{R} &= \rotb{A}{B}{R}, \eqnlabel{refrotidentA} \\
\rotb{B}{G}{R} \, \rotb{GA}{B}{R} &= \rotb{A}{G}{R}, & \rotb{G}{B}{R}\trans \, \rotb{GA}{B}{R} \, \rotb{G}{B}{R} &= \rotb{A}{B}{R}. \eqnlabel{refrotidentB}
\end{align}
Composition and inversion is given by
\begin{align}
\rotb{GA}{C}{R} &= \rotb{GB}{C}{R} \, \rotb{GA}{B}{R}, & \rotb{GA}{B}{R}\trans &= \rotb{GB}{A}{R}. \eqnlabel{refrotcompositioninversion}
\end{align}
Finally, a change of referenced frame is given by
\begin{equation}
\rotb{HA}{B}{R} = \rotb{H}{G}{R} \, \rotb{GA}{B}{R} \, \rotb{H}{G}{R}\trans. \eqnlabel{refrotrefchange}
\end{equation}
Referenced rotations are required for \secref{comp_mismatch_yaw_tilt}.

\section{Definition of Tilt Rotations}
\seclabel{defn_tilt_rotations}

In the fused angles and tilt angles representations, the yaw component of 
rotation is quantified using the \emph{fused yaw} \cite{Allgeuer2015a}, defined 
below in \secref{defn_fused_yaw}. Tilt rotations are exactly the rotations that 
have a zero fused yaw component. As such, all rotations can be neatly 
partitioned into their fused yaw and tilt rotation components, as described 
later in \secref{decomp_yaw_tilt}. Tilt rotations can however also be 
characterised as precisely the rotations that can be expressed as a pure 
rotation about a vector in the $\vect{xy}$ plane. The following sections detail 
the various available parameterisations of tilt rotations.

\subsection{Definition of Fused Yaw}
\seclabel{defn_fused_yaw}

Given a rotation from \fr{G} to \fr{B}, as illustrated in 
\figref{tilt_parameters}, we define a frame \fr{A} by rotating \fr{B} in such a 
way that $\vecs{z}{B}$ rotates onto $\vecs{z}{G}$ in the most direct way 
possible, within the plane that contains these two vectors. Note that this 
rotation of \fr{B} is in the opposite direction to the arrow labelled `$\alpha$' 
in the figure. The fused yaw $\psi \in (-\pi,\pi]$ is then given by the angle of 
the pure z-rotation from \fr{G} to \fr{A}. The remaining rotation from \fr{A} to 
\fr{B} is a tilt rotation, and is referred to as the \emph{tilt rotation 
component} of the total rotation. If $\quatb{G}{B}{q} = (w,x,y,z) \in \Q$ is the 
quaternion rotation from \fr{G} to \fr{B}, then mathematically the fused yaw is 
given by
\begin{equation}
\psi = \wrap\bigl(2\atantwo(z,w)\bigr), \eqnlabel{psidefn}
\end{equation}
where $\wrap(\cdot)$ is a function that wraps an angle to $(-\pi,\pi]$. The 
fused yaw has an essential discontinuity at $w = z = 0$, which is simply 
referred to as the fused yaw singularity.

\begin{figure}[!t]
\parbox{\linewidth}{\centering\includegraphics[width=1.0\linewidth]{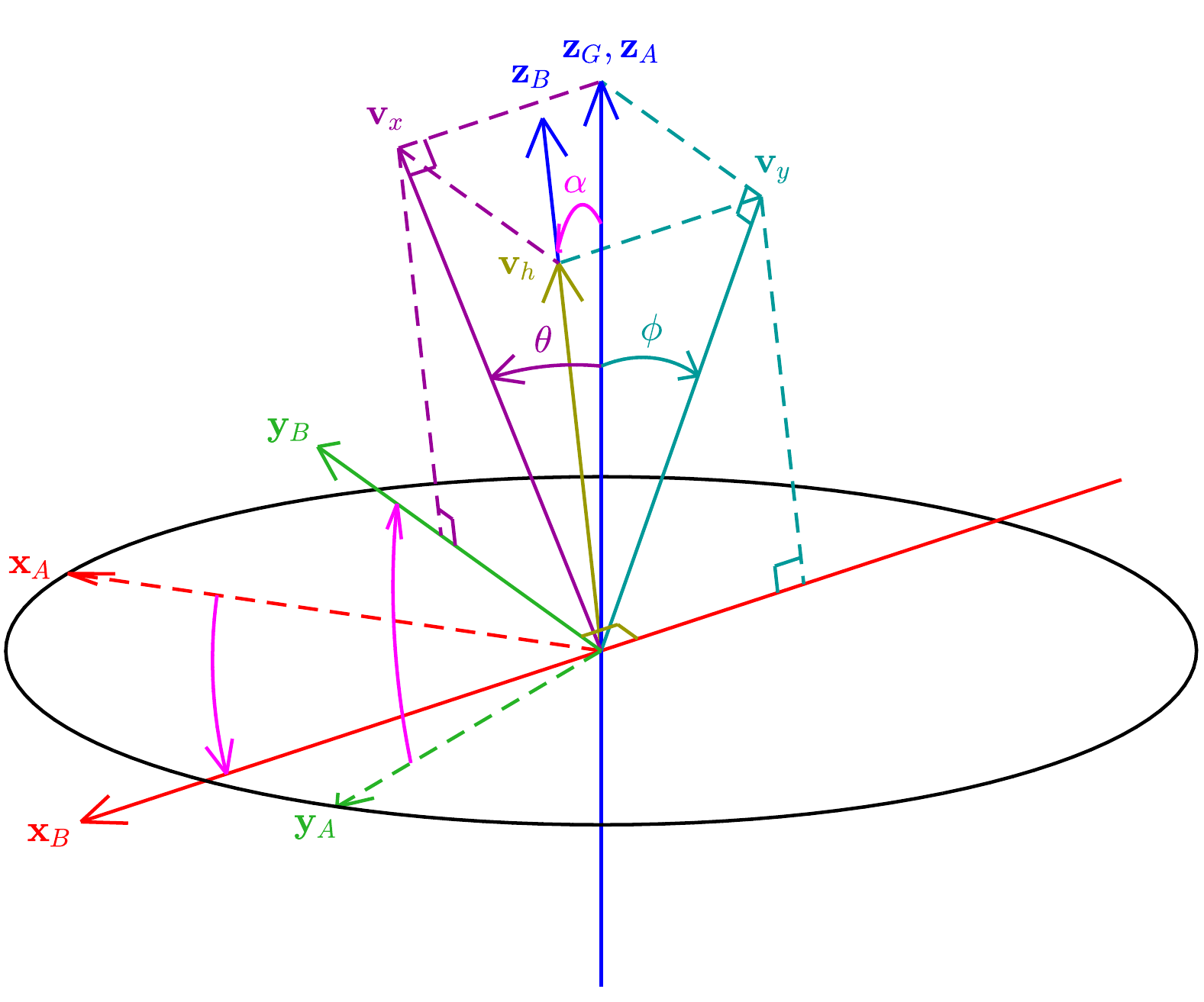}}
\caption{Diagram of the fused angles parameters $(\theta, \phi, h)$. The 
rotation by $\alpha$ from \fr{A} to \fr{B} is reparameterised by the angles 
$\theta$ and $\phi$, defined between $\vecs{z}{G}$ and the 
$\vecs{y}{B}\vecs{z}{B}$ and $\vecs{x}{B}\vecs{z}{B}$ planes respectively. The 
hemisphere $h$ is $1$ if $\vecs{v}{h}$ is parallel to $\vecs{z}{B}$, and $-1$ if 
it is antiparallel.}
\figlabel{fused_parameters}
\vspace{-2ex}
\end{figure}
\subsection{Parameterisations of Tilt Rotations}
\seclabel{params_tilt_rots}

\subsubsection{Tilt Angles Parameterisation}
\seclabel{param_tilt}

The tilt rotation component, i.e.\ the rotation from \fr{A} to \fr{B}, can be 
parameterised by the two parameters $(\gamma, \alpha)$. The \emph{tilt angle} 
$\alpha$ is the magnitude of the rotation, and the \emph{tilt axis angle} 
$\gamma$ defines the \emph{tilt axis} $\vect{\hat{v}}$ in the 
$\vecs{x}{G}\vecs{y}{G}$ plane about which the rotation occurs, as shown in 
\figref{tilt_parameters}. Together with the fused yaw $\psi$, these two 
parameters define the \emph{tilt angles} representation
\begin{equation}
\rotb{G\mspace{2mu}}{B}{T} = (\psi,\gamma,\alpha) \in (-\pi,\pi] {\times} (-\pi,\pi] {\times} [0,\pi] \defeq \T. \eqnlabel{tiltdefn}
\end{equation}
If $\rotb{G}{B}{R} \in \text{SO}(3)$ is the rotation matrix for the rotation 
from \fr{G} to \fr{B}, and $R_{ij}$ are the corresponding rotation matrix 
entries, then mathematically, $\gamma$ and $\alpha$ are given by
\begin{align}
\gamma &= \atantwo(-R_{31}, R_{32}), & \alpha &= \acos(R_{33}). \eqnlabel{gammaalphadefn}
\end{align}
By extending the domain of $\alpha$ to $[0,\infty)$, tilt rotations of 
arbitrary magnitude can be parameterised. This allows the magnitude of rotations 
greater than half a revolution to be captured, instead of just regarding the 
final resulting coordinate frame as a plain orientation.

\subsubsection{Fused Angles Parameterisation}
\seclabel{param_fused}

The tilt rotation component from \fr{A} to \fr{B} can also be parameterised 
using the three parameters $(\theta, \phi, h)$. The \emph{fused pitch} $\theta$ 
and \emph{fused roll} $\phi$ are defined as the signed angles between 
$\vecs{z}{G}$ and the $\vecs{y}{B}\vecs{z}{B}$ and $\vecs{x}{B}\vecs{z}{B}$ 
planes respectively, as shown in \figref{fused_parameters}. The 
\emph{hemisphere} $h \in \{-1,1\}$ is a binary parameter specifying whether 
$\vecs{z}{G}$ and $\vecs{z}{B}$ are in the same hemisphere or not. The complete 
\emph{fused angles} representation is then given by
\begin{equation}
\mspace{-8mu}\rotb{G\mspace{1mu}}{B}{\mspace{-2mu}F} \mspace{-5mu}=\mspace{-4mu} (\psi,\theta,\phi,h) \mspace{-4mu}\in\mspace{-4mu} (\!-\pi,\pi] {\times} [\mspace{-1mu}-\hpi,\hpi] {\times} [\mspace{-1mu}-\hpi,\hpi] {\times} \{\!-1,1\}\mspace{-1mu}.\mspace{-3mu} \eqnlabel{fuseddefn}
\end{equation}
Mathematically, the fused angles parameters are given by
\begin{align}
\mspace{-6mu}\theta &= \asin(-R_{31}), & \phi &= \asin(R_{32}), & h &= \sign(R_{33}). \eqnlabel{thetaphihdefn}
\end{align}
The relationship between $(\theta, \phi)$ and $(\gamma, \alpha)$ is given by
\begin{align}
\theta &= \asin(\sin\alpha\sin\gamma), & \phi &= \asin(\sin\alpha\cos\gamma). \eqnlabel{linkfusedtilt}
\end{align}

\subsubsection{Z-Vector Parameterisation}
\seclabel{param_zvector}

Recalling \eqnref{rotmatrowcolvec}, it can be seen from \eqnref{gammaalphadefn} 
and \eqnref{thetaphihdefn} that tilt rotations are a unique function of the 
corresponding \emph{z-vector} $\vecbs{B}{z}{G}$---at least away from the fused 
yaw singularity. In fact,
\begin{align}
\vecbs{B}{z}{G} &= \bigl( -\sin\alpha\sin\gamma,\, \sin\alpha\cos\gamma,\, \cos\alpha \bigr) \eqnlabel{BzGtilt} \\
&= \Bigl( -\sin\theta,\, \sin\phi,\, h\sqrt{1 - \sin^2\theta - \sin^2\phi} \, \Bigr)\mspace{-2mu}. \eqnlabel{BzGfused}
\end{align}

\subsubsection{Quaternion Parameterisation}
\seclabel{param_quat}

A rotation is a tilt rotation if and only if the z-component of the 
corresponding quaternion is zero. As such, a tilt rotation can be parameterised 
by its quaternion parameters
\begin{equation}
\mspace{-6mu}q = (w,x,y,0) = (\cos\tfrac{\alpha}{2},\, \sin\tfrac{\alpha}{2} \cos\gamma,\, \sin\tfrac{\alpha}{2} \sin\gamma,\, 0). \eqnlabel{quatfromgammaalpha}
\end{equation}

\section{Definition of the Tilt Phase Space}
\seclabel{defn_tilt_phase}

When working with tilt rotations and how to \emph{combine} them, the tilt angles 
parameterisation is frequently the representation of choice. In addition to 
providing an intuitive notion of the direction of tilt, which is helpful in many 
applications, the tilt angle parameter can also be extended to the domain 
$\alpha \in [0,\infty)$, allowing tilt rotations of unbounded magnitude to be 
represented and used for analysis. Rotation composition as a method of combining 
tilt rotations is unsuitable however, as the composition of tilt rotations is 
not closed, not commutative, and invariantly returns a bounded $\alpha$ for the 
rotation output. This shortcoming is one of the features that is addressed by 
the tilt phase space.

The tilt angles parameterisation also has the problem that it has a 
discontinuity in the tilt axis angle $\gamma$ at the identity rotation, and that 
the tilt angle $\alpha$ is not differentiable there due to a cusp 
\cite{Allgeuer2015a}. This leads to numerical and algorithmic difficulties, in 
particular if attempting to express tilt angles velocities, and attempting to 
relate them to angular velocities. This problem is similarly addressed by the 
tilt phase space.

There are two main variations of the tilt phase space, \emph{relative} and 
\emph{absolute}, and for each there is a choice of 2D or 3D, depending on 
whether the fused yaw is included or not. All four of these spaces are entirely 
analogous, however, so we can speak of just `the' tilt phase space.

\subsection{Relative Tilt Phase Space}
\seclabel{defn_rel_tilt_phase}

This is the default variant of the tilt phase space. By convention, the 
qualifier `relative' is only ever used if it is required to explicitly 
differentiate between the relative and absolute variants. Given the tilt angles 
representation $T = (\psi, \gamma, \alpha)$ of a rotation, the equivalent 
\emph{3D tilt phase space} representation is given by
\begin{equation}
P = (p_x, p_y, p_z) = (\alpha\cos\gamma,\, \alpha\sin\gamma,\, \psi) \in \R^3 \equiv \P^3. \eqnlabel{P3defn}
\end{equation}
Often when working with tilt rotations, the fused yaw component is not required. 
In such cases, the \emph{2D tilt phase space} representation can be used 
instead, given by
\begin{equation}
P = (p_x, p_y) = (\alpha\cos\gamma,\, \alpha\sin\gamma) \in \R^2 \equiv \P^2. \eqnlabel{P2defn}
\end{equation}
This is the predominant formulation of the tilt phase space that is used in 
analysis. Note that in \eqnrefs{P3defn}{P2defn}, a domain of $\R$ has been 
specified for the parameters, to naturally be able to represent rotations of 
more than 180\degree. The conversion from the relative tilt phase space to tilt 
angles is given by
\begin{align}
\psi &= p_z, & \gamma &= \atantwo(p_y, p_x), & \alpha &= \sqrt{p_x^2 + p_y^2}\mspace{2mu}. \eqnlabel{tiltfromphase}
\end{align}
The relationship between the 2D tilt phase parameters and the tilt angles 
parameters is shown visually in \figref{tilt_phase_defn}. The 2D tilt phase is 
to the tilt angles $(\gamma, \alpha)$ for tilt rotations, what cartesian 
coordinates is to polar coordinates for points.

It should be noted that the tilt phase parameters are globally continuous and 
smooth functions of the underlying rotation, in the sense that they are 
infinitely differentiable. This is a critically important property, and it is 
not trivial to see why this holds at $p_x = p_y = 0$, but it provably does. As 
such, differentiated quantities, such as for example tilt phase velocities and 
accelerations, are well-defined everywhere and can be smoothly converted to 
other representations, such as for example angular velocities and angular 
accelerations.

\subsection{Absolute Tilt Phase Space}
\seclabel{defn_abs_tilt_phase}

The absolute tilt phase space shares the same definition as the relative tilt 
phase space, only with the absolute tilt axis angle $\gammat = \gampsi$ being 
used instead of $\gamma$. That is,
\begin{alignat}{3}
\tilde{P} &= (\pxt, \pyt, \pzt) &&= (\alpha\cos\gammat,\, \alpha\sin\gammat,\, \psi) &&\in \R^3 \equiv \tilde{\P}^3, \eqnlabel{P3tildedefn} \\
\tilde{P} &= (\pxt, \pyt) &&= (\alpha\cos\gammat,\, \alpha\sin\gammat) &&\in \R^2 \equiv \tilde{\P}^2. \eqnlabel{P2tildedefn}
\end{alignat}
The conversions to the relative space and back are given by
\begin{equation}
\begin{alignedat}{4}
p_x &= &&c_\psi \pxt + s_\psi \pyt, && \pxt &&= c_\psi p_x - s_\psi p_y, \\
p_y &= -&&s_\psi \pxt + c_\psi \pyt, &\mspace{60mu}& \pyt &&= s_\psi p_x + c_\psi p_y,
\end{alignedat}
\eqnlabel{Ptildeconversion}
\end{equation}
where $c_\ast \equiv \cos\ast$ and $s_\ast \equiv \sin\ast$. As the two spaces 
are so tightly linked, all of the results that hold for one space correspond 
trivially to analogous results for the other. The two spaces are also negative 
inverses of one another, i.e.
\begin{align}
P\inv &= -\tilde{P}, & \tilde{P}\inv &= -P. \eqnlabel{Pinverses}
\end{align}
As $\tilde{P} \equiv P$ for pure tilt rotations, as a corollary
\begin{equation}
p_z = 0 \mspace{12mu} \implies \mspace{10mu} P\inv = -P.
\end{equation}

\begin{figure}[!t]
\parbox{\linewidth}{\centering\includegraphics[width=0.75\linewidth]{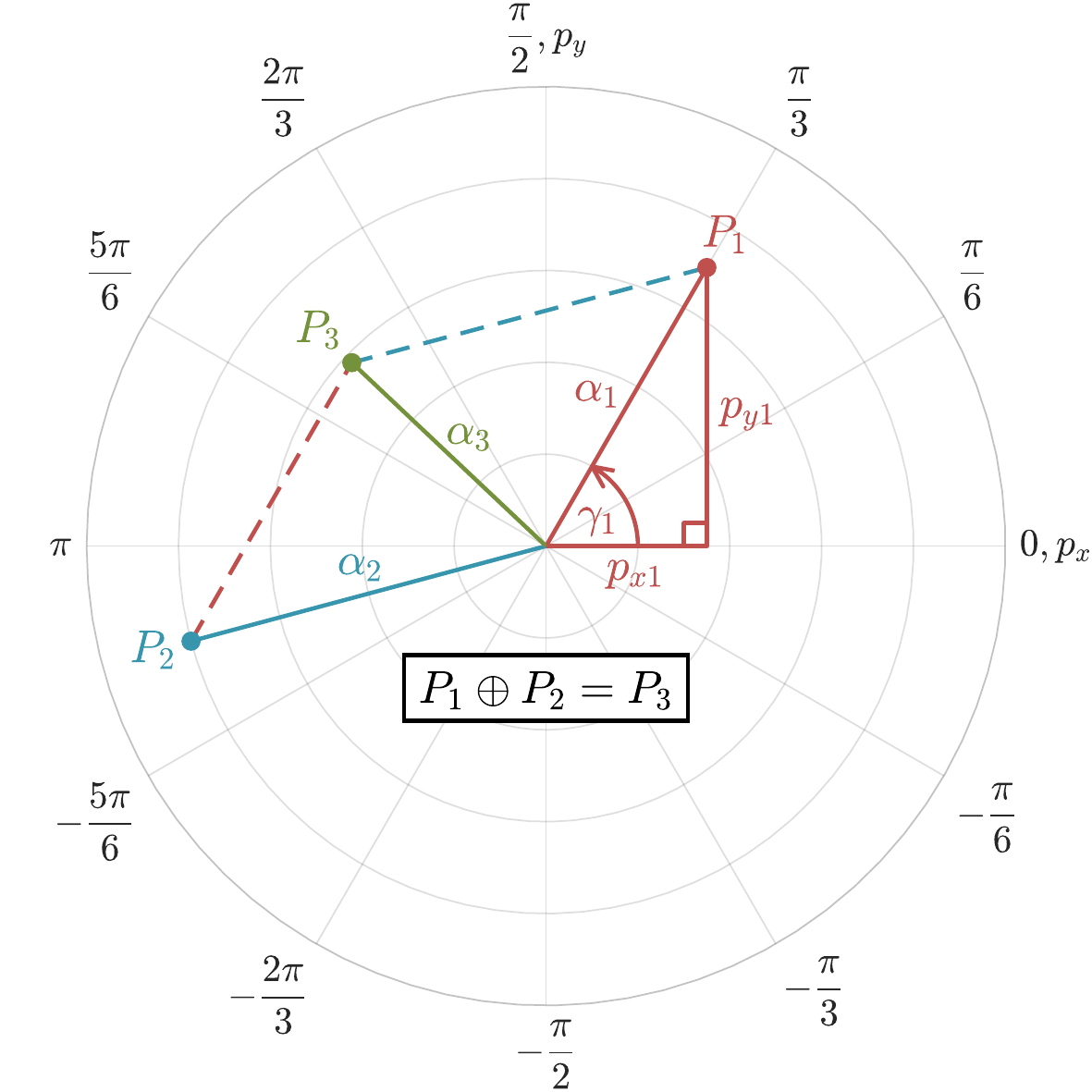}}
\caption{Tilt phase plot demonstrating tilt vector addition (see 
\secref{tilt_vector_addition}), and the link between the tilt phase and tilt 
angles parameters.}
\figlabel{tilt_phase_defn}
\vspace{-2ex}
\end{figure}
\subsection{Fundamental Properties of the Tilt Phase Space}
\seclabel{fundamental_properties}

Similar to fused angles, the tilt phase space has numerous properties 
\cite{Allgeuer2018a} that set it apart from alternatives such as for example 
Euler angles. The parameters are mutually independent and correspond to 
different major planes of rotation, the yaw, i.e.\ $p_z$, is axisymmetric, and 
the two remaining tilt parameters are axisymmetric and correspond to each other 
in behaviour. A unique property of the tilt phase space however, is 
\emph{magnitude axisymmetry}. This relates to the fact that equal angle 
magnitude rotations in any tilt direction have equal norms in the 2D tilt phase 
space. This is not the case for the pitch and roll components of fused angles.

\subsection{Tilt Vector Addition}
\seclabel{tilt_vector_addition}

Although the composition of tilt rotations is not commutative and in general 
does not produce a tilt rotation as an output, the 2D tilt phase space provides 
a way of defining a useful and meaningful addition operator for tilt rotations 
that is closed, commutative and associative. This is referred to as tilt vector 
addition, and for $P_1, P_2 \in \P^2$ is given by
\begin{equation}
\begin{split}
P_3 = P_1 \oplus P_2 &= (p_{x1} + p_{x2},\, p_{y1} + p_{y2}) \in \P^2 \\
&= (\alpha_1 c_{\gamma_1} + \alpha_2 c_{\gamma_2},\, \alpha_1 s_{\gamma_1} + \alpha_2 s_{\gamma_2}).
\end{split}
\eqnlabel{tiltvectoraddition}
\end{equation}
\eqnref{tiltfromphase} can be used to calculate the tilt angles parameters 
$(\gamma_3, \alpha_3)$ corresponding to $P_3$. In abbreviated form, we write
\begin{equation}
(\gamma_1, \alpha_1) \oplus (\gamma_2, \alpha_2) = (\gamma_3, \alpha_3). \eqnlabel{gammaalphaaddition}
\end{equation}
The action of tilt vector addition is illustrated in \figref{tilt_phase_defn}. 
Completely analogous definitions of tilt vector addition hold for the absolute 
tilt phase space, where it should be noted that the addition of absolute phases 
is consistent with the addition of relative phases, as long as it is being done 
for a fixed fused yaw. In other words, if any two tilt rotations $(\gamma, 
\alpha)$ are converted to both relative and absolute tilt phases, and added in 
each representation, then the outputs expressed as, for example, quaternions are 
identical.

\subsection{Vector Space of Tilt Rotations}
\seclabel{vector_space}

Based on \eqnref{tiltvectoraddition}, it is easy to see that $(\P^2, \oplus)$ is 
an abelian group. For $\lambda \in \R$, we define scalar multiplication as
\begin{equation}
\lambda P = (\lambda p_x, \lambda p_y).
\end{equation}
This completes $\P^2$ as a vector space over $\R$ that is isomorphic to $\R^2$, 
as suggestively written in \eqnref{P2defn}. This is referred to as the 
\emph{vector space of tilt rotations}. The vector space of absolute tilt 
rotations is similarly defined. It is easy to verify that the additive inverses 
in these vector spaces in fact correspond to the true inverses of the 
corresponding tilt rotations. The identity vector, i.e.\ the zero vector, also 
corresponds to the true identity tilt rotation.

Given that tilt rotations can be formulated as a vector space, many useful 
properties, results, concepts and algorithms come for `free'. For instance, 
combining tilt vector addition with scalar multiplication leads to a trivial 
definition of the mean of a set of tilt rotations. Other useful corollaries of 
having a vector space, that are however outside of the scope of this paper, 
include results involving differentiation, integration, metrics and general 
linear algebra.

Having a vector space of tilt rotations also allows cubic spline interpolation 
to be performed, with the guarantee that all produced intermediate rotations are 
indeed tilt rotations. Other methods of orientation spline 
interpolation---involving for example the logarithmic and exponential map and 
working with the Lie algebra $\mathfrak{so}(3)$---in general do not have this 
required property, are significantly more computationally expensive, and cannot 
deal with rotations above 180\degree, albeit for the benefit of often being 
bi-invariant \cite{Kang1999}. In the inherently asymmetrical situation where 
there is a clear notion of `up', however, bi-invariance is not seen as a 
decisive advantage, especially when observing that tilt phase space cubic spline 
interpolation is in fact invariant about the `up' z-axis. It should be noted 
that the optimal minimum angular acceleration interpolating curve between 
orientations is an involved three-dimensional, fourth-order nonlinear two-point 
boundary value problem, does not in general admit a closed form solution 
\cite{Kang1999}, and is thus frequently not an option.

\section{Properties of Tilt Rotations and\texorpdfstring{\\}{ }the Tilt Phase Space}
\seclabel{properties}

In this section, various results pertaining to tilt rotations and the tilt phase 
space are explored and presented.

\subsection{Spherical Linear Interpolation Between Tilt Rotations}
\seclabel{slerp}

Spherical linear interpolation (slerp) is a way of interpolating rotations that 
is torque-minimal and constant angular velocity. Given $q_0, q_1 \in \Q$ and $u 
\in [0,1]$, slerp is given by
\begin{equation}
\slerp(q_0, q_1, u) = \left[ \frac{\sin((1-u)\Omega)}{\sin{\Omega}} \right] \! q_0 + \left[ \frac{\sin u\Omega}{\sin{\Omega}} \right] \! q_1, \eqnlabel{slerpsumdefn}
\end{equation}
where $\Omega = \acos(q_0 \dotp q_1)$, and $q_0$, $q_1$ need to be mutually in 
the same 4D hemisphere. It can be demonstrated that slerp between any two tilt 
rotations always produces a tilt rotation---a property that is less trivial than 
it may seem. As such, tilt rotations can always cleanly and easily be 
interpolated, without affecting the fused yaw. Furthermore, for any $\hat{q} \in 
\Q$,
\begin{equation}
\slerp(\hat{q}q_0, \hat{q}q_1, u) = \hat{q} \slerp(q_0,q_1,u). \eqnlabel{slerpinvariant}
\end{equation}
Thus, by choosing $\hat{q} = q_z(\psi)$, i.e.\ the quaternion corresponding to a 
z-rotation by $\psi$, it follows that slerp between any two rotations of equal 
fused yaw always produces an output rotation of exactly the same fused yaw. That 
is,
\begin{equation}
\yawof{q_0} = \yawof{q_1} = \psi \implies \yawof{\slerp(q_0, q_1, u)} = \psi, \eqnlabel{slerpequalpsi}
\end{equation}
where the function $\Psi(\cdot)$ returns the fused yaw of a rotation.

\subsection{Relation Between Fused Angles and the Tilt Phase Space}
\seclabel{relation_fused_tilt_phase}

Just like the fused roll $\phi$ and fused pitch $\theta$ can be seen to quantify 
the amount of tilt rotation about the x and y-axes respectively, the tilt phase 
space parameters $(p_x, p_y)$ also do exactly that. From \eqnref{tiltfromphase}, 
a tilt rotation of pure $p_x$ corresponds by definition to a $\gamma$ of 0 or 
$\pi$, and hence corresponds to a pure x-rotation, as the tilt axis 
$\vect{\hat{v}}$ (see \figref{tilt_parameters}) is then parallel or antiparallel 
to the x-axis. Similarly, a tilt rotation of pure $p_y$ corresponds to a 
$\gamma$ of $\pm\hpi$, and hence corresponds to a pure y-rotation. Furthermore, 
from \eqnref{linkfusedtilt},
\begin{align}
\sin\phi &= \sin\alpha \cos\gamma, & \sin\theta &= \sin\alpha \sin\gamma.
\end{align}
Applying the small angle approximation to $\phi$, $\theta$ and $\alpha$ yields
\begin{align}
\phi &\approx \alpha\cos\gamma = p_x, & \theta &\approx \alpha\sin\gamma = p_y.
\end{align}
In fact, $p_x$ and $p_y$ are from a mathematical perspective the 
$\mathcal{O}(\alpha^3)$ Taylor series approximations of $\phi$ and $\theta$, 
respectively. As such, the 2D tilt phase parameters mimic fused angles for tilt 
rotations of small magnitudes, but increase linearly to infinity for large 
magnitudes. This is unlike $\phi$ and $\theta$, which loop around to correctly 
represent the resulting orientation. For example, $(\gamma, \pi)$ and $(\gamma, 
-\pi)$ both correspond to the same orientation in terms of fused angles, but are 
completely different tilt phase rotations. Put concisely, the tilt phase space 
is for unbounded tilt rotations, i.e.\ tilt rotations of more than 180\degree, 
what fused angles is for bounded tilt rotations---a way of concurrently 
quantifying in an axisymmetric manner the amount of rotation about each of the 
coordinate axes.

The relative differences between the fused angles and tilt phase space 
parameters are plotted in \figref{fused_phase_comparison} as a function of $p_x$ 
and $p_y$. The plotted values are expressed as ratios of the tilt rotation 
magnitude $\alpha$. It can be observed that the errors in the approximations 
$p_x \approx \phi$ and $p_y \approx \theta$ are in fact notably less than the 
error involved in assuming $\sin\alpha \approx \alpha$, showing how close the 
parameters are for even medium rotations.

\subsection{Interpretation of Tilt Vector Addition}
\seclabel{interpretation_tilt_addition}

The definition used for tilt vector addition is in part motivated by the 
unambiguous commutative way in which angular velocities, unlike angular 
rotations, can be added. The interpretation thereof as the 
addition of instantaneous angular velocities extends naturally to unbounded tilt 
rotations, as angular velocities can also have arbitrary magnitude, and do not 
`wrap around' like rotations do. Suppose we wish to add two tilt rotations $(\gamma_1, \alpha_1)$ 
and $(\gamma_2, \alpha_2)$, as in \eqnref{gammaalphaaddition}. The tilt rotation 
given by $(\gamma_1, \alpha_1)$ is equivalent to applying an angular velocity of 
$\vecbs{G}{\Omega}{1}$ for $t$ seconds, where
\begin{equation}
\vecbs{G}{\Omega}{1} = \tfrac{1}{t} (\alpha_1 c_{\gamma_1},\, \alpha_1 s_{\gamma_1},\, 0).
\end{equation}
Similarly, $(\gamma_2, \alpha_2)$ is equivalent to $t$ seconds of
\begin{equation}
\vecbs{G}{\Omega}{2} = \tfrac{1}{t} (\alpha_2 c_{\gamma_2},\, \alpha_2 s_{\gamma_2},\, 0).
\end{equation}
Thus, if both angular velocities are applied at the same time, the resulting 
total angular velocity is given by $\vecbs{G}{\Omega}{3}$, where
\begin{equation}
\begin{split}
\vecbs{G}{\Omega}{3} &= \vecbs{G}{\Omega}{1} + \vecbs{G}{\Omega}{2} \\
&= \tfrac{1}{t} (\alpha_1 c_{\gamma_1} + \alpha_2 c_{\gamma_2},\, \alpha_1 s_{\gamma_1} + \alpha_2 s_{\gamma_2},\, 0).
\end{split}
\end{equation}
Given that $\vecbs{G}{\Omega}{3}$ is then applied for $t$ seconds, the parallel 
to the definition \eqnref{tiltvectoraddition} of tilt vector addition is clear.

\begin{figure}[!t]
\parbox{\linewidth}{\centering\includegraphics[width=1.0\linewidth]{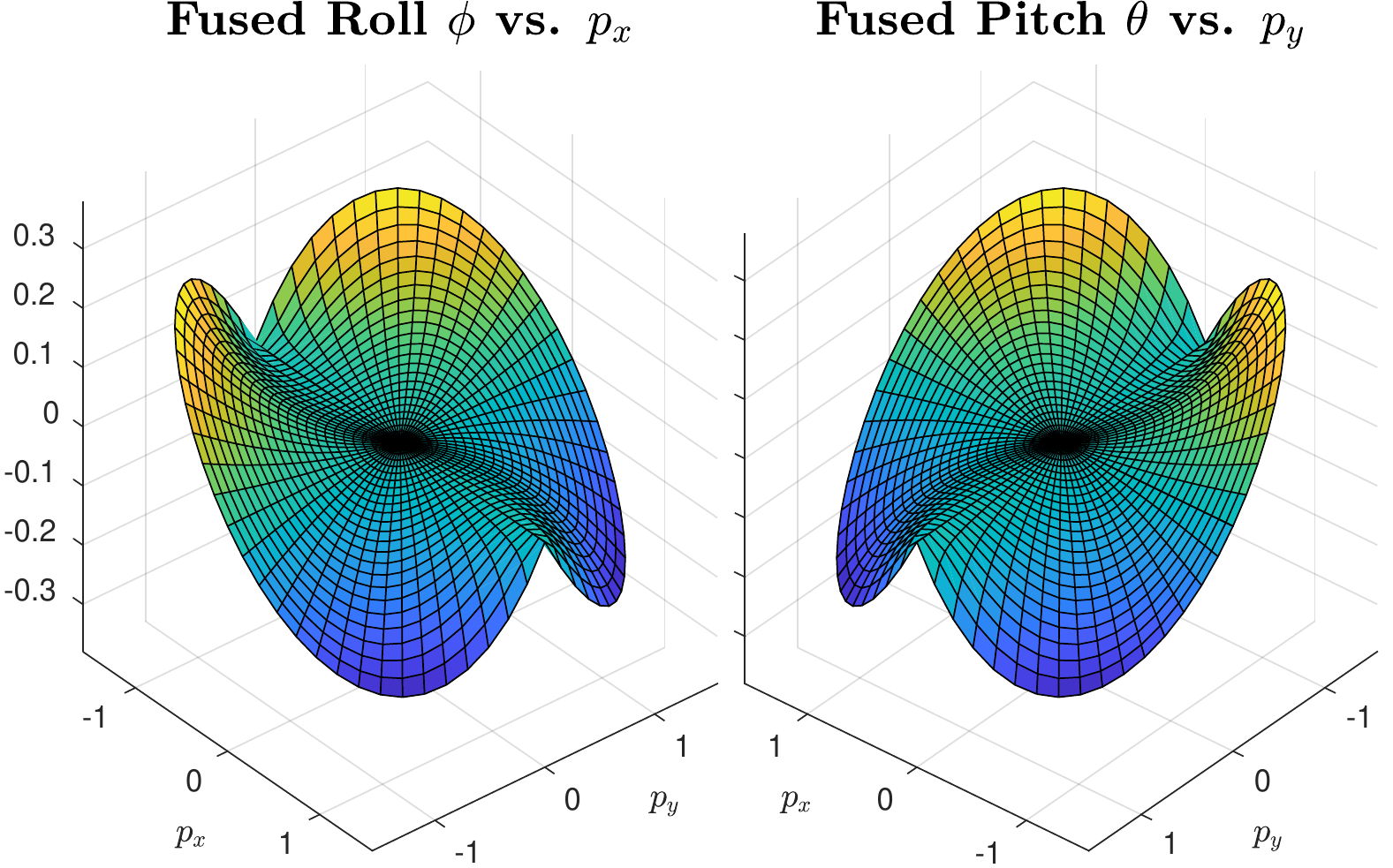}}
\caption{Plots of the relative difference, as a ratio of the tilt rotation 
magnitude $\alpha$, between the fused angles and tilt phase space parameters. 
For $\alpha = 1 \approx 57.3\degreem$, the maximum relative difference is just 
7.1\%, compared to 15.9\% for the equivalent small angle approximation 
$\sin\alpha \approx \alpha$. For $\alpha = \hpi = 90\degreem$, these numbers are 
21.1\% and 36.3\%, respectively.}
\figlabel{fused_phase_comparison}
\vspace{-2ex}
\end{figure}
\subsection{Rotational Velocities}
\seclabel{rotational_velocities}

When working with trajectories in either the 2D tilt rotation space or the full 
3D rotation space, like for example for cubic spline interpolation, it is 
necessary to be able to relate rotational velocities between the various 
representations. In particular, the relationships between the tilt phase 
velocities $\dot{P} = \bigl( \pxd, \pyd, \pzd \bigr)$ and $\dot{\tilde{P}} = 
\bigl( \pxtd, \pytd, \pztd \bigr)$, tilt angles velocities $\dot{T} = \bigl( 
\dot{\psi}, \dot{\gamma}, \dot{\alpha} \bigr)$, absolute tilt axis angle 
velocity $\dot{\gammat}$, and angular velocity $\vecb{G}{\Omega}$, are of great 
interest, and are thus presented here. All velocity conversions are shown for $T 
= (\psi, \gamma, \alpha)$.

To begin, we note that $\gammat = \gampsi$ and $\psi = p_z = \pzt$, so
\begin{align}
\dot{\gammat} &= \dot{\gamma} + \dot{\psi}, & \dot{\psi} &= \pzd = \pztd. \eqnlabel{gammapsipzdotequiv}
\end{align}
As a result of the latter equation, for conversions between $\dot{T}$, $\dot{P}$ 
and $\dot{\tilde{P}}$, we only need to focus on the x and y-components.

\subsubsection{Tilt Phase Velocity Conversions}
\seclabel{velconv_phase}

Keeping \eqnref{gammapsipzdotequiv} in mind, the conversion from the relative 
tilt phase velocity $\dot{P}$ to the absolute tilt phase velocity 
$\dot{\tilde{P}}$ is given by
\begin{equation}
\begin{aligned}
\pxtd &= c_\psi \pxd - s_\psi \pyd - \pyt \pzd, \\
\pytd &= s_\psi \pxd + c_\psi \pyd + \pxt \pzd.
\end{aligned} \eqnlabel{ptvelfrompvel}
\end{equation}
The conversion from absolute back to relative is given by
\begin{equation}
\begin{alignedat}{2}
\pxd &=  &&c_\psi \pxtd + s_\psi \pytd + p_y \pztd, \\
\pyd &= -&&s_\psi \pxtd + c_\psi \pytd - p_x \pztd.
\end{alignedat} \eqnlabel{pvelfromptvel}
\end{equation}

\subsubsection{Tilt Phase Velocity \texorpdfstring{$\leftrightarrow$}{<->} Tilt Angles Velocity}
\seclabel{velconv_phase_tilt}

Together with \eqnref{gammapsipzdotequiv}, the conversion from the tilt phase 
velocities $\dot{P}$, $\dot{\tilde{P}}$ to the tilt angles velocity $\dot{T}$ is 
given by
\begin{align}
\dot{\gamma} &= \tfrac{1}{\alpha} \bigl( c_\gamma \pyd - s_\gamma \pxd \bigr), & \dot{\gammat} &= \tfrac{1}{\alpha} \bigl( c_\gammat \pytd - s_\gammat \pxtd \bigr), \eqnlabel{gammavelfrompvel} \\
\dot{\alpha} &= c_\gamma \pxd + s_\gamma \pyd, & \dot{\alpha} &= c_\gammat \pxtd + s_\gammat \pytd. \eqnlabel{alphavelfrompvel}
\end{align}
We note that $\gamma$, $\gammat$ have essential singularities at $\alpha = 0$, 
so as expected, the velocities in \eqnref{gammavelfrompvel} are infinite in this 
case. The reverse conversions are always stable, however, and given by
\begin{equation}
\begin{aligned}
\pxd &= c_\gamma \dot{\alpha} - s_\gamma \alpha \dot{\gamma}, \mspace{24mu}&\mspace{24mu} \pxtd &= c_\gammat \dot{\alpha} - s_\gammat \alpha \dot{\gammat}, \\
\pyd &= s_\gamma \dot{\alpha} + c_\gamma \alpha \dot{\gamma}, \mspace{24mu}&\mspace{24mu} \pytd &= s_\gammat \dot{\alpha} + c_\gammat \alpha \dot{\gammat}.
\end{aligned} \eqnlabel{pvelfromtiltvel}
\end{equation}

\subsubsection{Tilt Angles Velocity \texorpdfstring{$\leftrightarrow$}{<->} Angular Velocity}
\seclabel{velconv_tilt_angvel}

Being able to convert a rotational velocity to an angular velocity is important, 
for example, if one wishes to convert a 6D inverse kinematics velocity to a joint 
space velocity. For the tilt angles velocity $\dot{T}$, the conversion is given by
\begin{equation}
\vecb{G}{\Omega} = \bigl( c_\gammat \dot{\alpha} - s_\gammat s_\alpha \dot{\gamma},\, s_\gammat \dot{\alpha} + c_\gammat s_\alpha \dot{\gamma},\, \dot{\psi} + (1 - c_\alpha) \dot{\gamma} \bigr). \eqnlabel{angvelfromtiltvel}
\end{equation}
The reverse conversion from $\vecb{G}{\Omega}$ to $\dot{T}$ is given by
\begin{align}
&\mspace{-24mu}
\begin{aligned}
\dot{\psi} &= \vecb{G}{\Omega} \dotp \vecs{v}{\psi}, &\mspace{36mu} \dot{\gamma} &= \vecb{G}{\Omega} \dotp \vecs{v}{\gamma}, \\
\dot{\alpha} &= \vecb{G}{\Omega} \dotp \vecs{v}{\alpha}, &\mspace{36mu} \dot{\gammat} &= \vecb{G}{\Omega} \dotp \vecs{v}{\gammat},
\eqnlabel{tiltvelfromangvel}
\end{aligned} \\
\vecs{v}{\psi} &= \tfrac{1}{1 + c_\alpha} \bigl( s_\alpha s_\gammat,\, -s_\alpha c_\gammat,\, 1 + c_\alpha \bigr), \eqnlabel{tiltvelfromangvelvpsi} \\
\vecs{v}{\gammat} &= \tfrac{1}{s_\alpha} \bigl( -c_\alpha s_\gammat,\, c_\alpha c_\gammat,\, s_\alpha \bigr), \eqnlabel{tiltvelfromangvelvgammat} \\
\vecs{v}{\gamma} &= \tfrac{1}{s_\alpha} \bigl( -s_\gammat,\, c_\gammat,\, 0 \bigr), \eqnlabel{tiltvelfromangvelvgamma} \\
\vecs{v}{\alpha} &= \bigl( c_\gammat,\, s_\gammat,\, 0 \bigr). \eqnlabel{tiltvelfromangvelvalpha}
\end{align}
Geometrically, it can be seen that $\vecs{v}{\psi}$ is along the angle bisector 
of $\vecs{z}{G}$ and $\vecs{z}{B}$---the two vectors that define the plane of the 
tilt rotation component (see \figref{tilt_parameters}). The vector 
$\vecs{v}{\alpha}$ corresponds geometrically to the unit normal vector of this 
plane, as expected from the definition of $\alpha$, and is in fact orthogonal to 
all three of the remaining vectors. The expressions for $\vecs{v}{\psi}$, 
$\vecs{v}{\gammat}$ and $\vecs{v}{\gamma}$ are valid away from $\alpha = \pi$, 
as the fused yaw $\psi$ has an unavoidable singularity there. The latter two 
also have problems when $\alpha = 0$, due to the tilt axis angle singularity. 
This shortcoming is addressed in the context of the tilt phase space in 
\secref{velconv_phase_angvel} below.

If we consider a rotation trajectory in the 2D tilt rotation space, it is clear 
that at every point on the trajectory we must have $\psi = 0$ and $\dot{\psi} = 
0$. From \eqnrefs{tiltvelfromangvel}{tiltvelfromangvelvpsi}, and the quaternion 
expression $q = (w, x, y, 0)$ given in \eqnref{quatfromgammaalpha}, we can 
deduce that at every point on the trajectory we must have
\begin{equation}
\vecb{G}{\Omega} \dotp (y,-x,w) = 0.
\end{equation}
This equation characterises what can be considered to be the tangent space to 
the differentiable manifold of tilt rotations.

\subsubsection{Tilt Phase Velocity \texorpdfstring{$\leftrightarrow$}{<->} Angular Velocity}
\seclabel{velconv_phase_angvel}

When converting between tilt phase velocities and angular velocities, it is 
natural to convert via tilt angles velocities. This causes unnecessary problems 
with the tilt axis angle singularity however, as neither the tilt phase 
representation nor the angular velocity actually has a singularity at $\alpha = 
0$, but the tilt angles representation does. By combining the required 
conversion equations, and taking care of the resulting removable singularity at 
$\alpha = 0$, robust conversions between tilt phase velocities and angular 
velocities can be achieved.

Together with \eqnref{gammapsipzdotequiv} and \eqnref{alphavelfrompvel}, the 
conversion from the tilt phase velocities $\dot{P}$, $\dot{\tilde{P}}$ to an 
angular velocity $\vecb{G}{\Omega}$ is given by
\begin{gather}
\vecb{G}{\Omega} = \bigl( c_\gammat \dot{\alpha} - S s_\gammat \dot{\gamma}_\alpha,\, s_\gammat \dot{\alpha} + S c_\gammat \dot{\gamma}_\alpha,\, \dot{\psi} + C \dot{\gamma}_\alpha \bigr), \eqnlabel{angvelfrompvel} \\
\dot{\gamma}_\alpha \equiv \alpha \dot{\gamma} = c_\gamma \pyd - s_\gamma \pxd = c_\gammat \pytd - s_\gammat \pxtd - \alpha \pztd, \eqnlabel{angvelfrompvelagamma} \\
\begin{aligned}
S &=
\begin{cases}
\tfrac{s_\alpha}{\alpha}, & \text{if $\alpha \neq 0$,} \\
1, & \text{if $\alpha = 0$,}
\end{cases} \mspace{18mu}&
C &=
\begin{cases}
\tfrac{1 - c_\alpha}{\alpha}, & \text{if $\alpha \neq 0$,} \\
0, & \text{if $\alpha = 0$.}
\end{cases}\mspace{-18mu}
\end{aligned}
\eqnlabel{angvelfrompvelSC}
\end{gather}
S and C are smooth functions of $\alpha$, as the removable singularity at the 
origin is in each case remedied by manual definition. This leads to a globally 
robust and smooth expression for $\vecb{G}{\Omega}$, including for $\alpha = 0$. 
In fact,
\begin{equation}
\alpha = 0 \implies \vecb{G}{\Omega} = \dot{\tilde{P}}. \eqnlabel{angvelptvelequal}
\end{equation}
This result exemplifies the link between angular velocities and the tilt phase 
space, as was discussed in the context of tilt vector addition in 
\secref{interpretation_tilt_addition}. The conversion from $\vecb{G}{\Omega}$ to 
the corresponding tilt phase velocities is given by
\begin{align}
&\mspace{-60mu}
\begin{aligned}
\pxd &= \vecb{G}{\Omega} \dotp \vecs{v}{x}, \mspace{9mu}& \pyd &= \vecb{G}{\Omega} \dotp \vecs{v}{y}, \mspace{9mu}& \pzd &= \vecb{G}{\Omega} \dotp \vecs{v}{\psi}, \eqnlabel{pvelfromangvel}
\end{aligned} \mspace{-36mu} \\
\vecs{v}{x} &= \bigl( c_\gamma c_\gammat + \tfrac{1}{S} s_\gamma s_\gammat,\, c_\gamma s_\gammat - \tfrac{1}{S} s_\gamma c_\gammat,\, 0 \bigr), \eqnlabel{pvelfromangvelvx} \\
\vecs{v}{y} &= \bigl( s_\gamma c_\gammat - \tfrac{1}{S} c_\gamma s_\gammat,\, s_\gamma s_\gammat + \tfrac{1}{S} c_\gamma c_\gammat,\, 0 \bigr). \eqnlabel{pvelfromangvelvy}
\end{align}
The corresponding equations for $\dot{\tilde{P}}$ are given by
\begin{align}
&\mspace{-60mu}
\begin{aligned}
\pxtd &= \vecb{G}{\Omega} \dotp \vecs{v}{\tilde{x}}, \mspace{9mu}& \pytd &= \vecb{G}{\Omega} \dotp \vecs{v}{\tilde{y}}, \mspace{9mu}& \pztd &= \vecb{G}{\Omega} \dotp \vecs{v}{\psi}, \eqnlabel{ptvelfromangvel}
\end{aligned} \mspace{-36mu} \\
\vecs{v}{\tilde{x}} &= \bigl( c_\gammat^2 + \tfrac{c_\alpha}{S} s_\gammat^2,\, c_\gammat s_\gammat (1 - \tfrac{c_\alpha}{S}),\, -\alpha s_\gammat \bigr), \eqnlabel{ptvelfromangvelvxt} \\
\vecs{v}{\tilde{y}} &= \bigl( c_\gammat s_\gammat (1 - \tfrac{c_\alpha}{S}),\, s_\gammat^2 + \tfrac{c_\alpha}{S} c_\gammat^2,\, \alpha c_\gammat \bigr). \eqnlabel{ptvelfromangvelvyt}
\end{align}
We note that $S \neq 0$ away from the fused yaw singularity, so $\tfrac{1}{S}$ 
is also smooth on this domain, and, critically, there is no cusp or irregularity 
at $\alpha = 0$. This is expected, given \eqnref{angvelptvelequal}.

\subsection{Rotation Decomposition into Yaw and Tilt}
\seclabel{decomp_yaw_tilt}

As mentioned previously, all rotations can be decomposed into their fused yaw 
and tilt rotation components, where these two components are then completely 
independent entities. For the tilt angles rotation $T = (\psi, \gamma, \alpha)$, 
the fused angles rotation $F = (\psi, \theta, \phi, h)$, and the quaternion $q$, 
we have
\begin{alignat}{2}
T &= T_y \circ T_t &&= T(\psi,0,0) \circ T(0,\gamma,\alpha), \eqnlabel{decomposetilt}\\
F &= F_y \circ F_t &&= F(\psi,0,0,1) \circ F(0,\theta,\phi,h), \eqnlabel{decomposefused}\\
q &= q_y q_t &&= (c_{\bar{\psi}}, 0, 0, s_{\bar{\psi}}) (c_{\bar{\alpha}},\, s_{\bar{\alpha}} c_\gamma,\, s_{\bar{\alpha}} s_\gamma,\, 0), \eqnlabel{decomposequat}
\end{alignat}
where `$\circ$' represents rotation composition, $\bar{\psi} \equiv \half\psi$, 
and ${\bar{\alpha} \equiv \half\alpha}$. For the rotation matrix R, we have
\begin{align}
\mspace{-9mu}R &= \rots{R}{y} \rots{R}{t}, \notag \\
&= \mspace{-5mu}
\setlength{\arraycolsep}{2.5pt}
\begin{bmatrix}
c_\psi & -s_\psi & 0 \\
s_\psi & c_\psi & 0 \\
0 & 0 & 1
\end{bmatrix}\mspace{-12mu}\begin{bmatrix}
c_\gamma^2 + c_\alpha s_\gamma^2 & c_\gamma s_\gamma (1 \!-\! c_\alpha) & s_\alpha s_\gamma \\
c_\gamma s_\gamma (1 \!-\! c_\alpha) & s_\gamma^2 + c_\alpha c_\gamma^2 & -s_\alpha c_\gamma \\
-s_\alpha s_\gamma & s_\alpha c_\gamma & c_\alpha
\end{bmatrix}\mspace{-8mu}.\mspace{-9mu} \eqnlabel{decomposerotmat}
\end{align}
By definition, $T_y$, $F_y$, $q_y$ and $R_y$ are all pure z-rotations, i.e.
\begin{equation}
\begin{aligned}
T_y &= T_z(\psi), &\mspace{48mu} F_y &= F_z(\psi), \\
q_y &= q_z(\psi), &\mspace{48mu} R_y &= R_z(\psi),
\end{aligned}
\end{equation}
where for example $R_z(\psi)$ is the rotation about the z-axis by $\psi$. $T_t$, 
$F_t$, $q_t$ and $R_t$ all quantify and parameterise the tilt rotation component 
of the respective full rotations.

\subsection{Rotation Composition from Yaw and Tilt}
\seclabel{comp_yaw_tilt}

Given that the fused yaw and tilt rotation component cleanly partition a 3D 
rotation, it is desired to be able to compose two arbitrary such components to 
construct a full rotation. This is trivial, using 
\eqnrefs{decomposetilt}{decomposequat}, if the required tilt rotation component 
is specified in the tilt angles, fused angles and/or quaternion 
parameterisations. If the tilt rotation component is specified in terms of the 
z-vector $\vecbs{B}{z}{G}$, then \eqnref{gammaalphadefn} and 
\eqnref{thetaphihdefn} can be used to generate the required tilt angles and 
fused angles representations, respectively. If the quaternion or rotation matrix 
representation is desired, the most direct and numerically safe method, however, 
is to directly construct the quaternion as follows.

The w and z-components are first calculated using
\begin{align}
N_{wz} &= \half(1 + \compbs{B}{z}{Gz}),
&\begin{aligned}
w &= c_{\bar{\psi}} \sqrt{N_{wz}}\mspace{1mu}, \\
z &= s_{\bar{\psi}} \sqrt{N_{wz}}\mspace{1mu}.
\end{aligned}
\end{align}
The x and y-components are then calculated as
\begin{gather}
\begin{split}
\tilde{x} &= \compbs{B}{z}{Gx} z + \compbs{B}{z}{Gy} w, \\
\tilde{y} &= \compbs{B}{z}{Gy} z - \compbs{B}{z}{Gx} w,
\end{split} \\
\begin{aligned}
A &= \sqrt{\frac{1 - N_{wz}}{\tilde{x}^2 + \tilde{y}^2}},
&\mspace{18mu}\begin{aligned}
x &= A \tilde{x}, \\
y &= A \tilde{y}.
\end{aligned}
\end{aligned} \eqnlabel{psiBzGquatxy}
\end{gather}
Being careful of the $\psi$ singularity, the final quaternion is then
\begin{equation}
q =
\begin{cases}
(0,1,0,0), & \text{if $\tilde{x} = \tilde{y} = 0$,} \\
(w,x,y,z), & \text{otherwise.}
\end{cases} \eqnlabel{psiBzGquatconstruct}
\end{equation}
For the special case that $\psi = 0$, we have
\begin{equation}
\begin{aligned}
w &= \sqrt{N_{wz}}, &\mspace{36mu} \tilde{x} &= \compbs{B}{z}{Gy}, \\
z &= 0, & \tilde{y} &= -\compbs{B}{z}{Gx}.
\end{aligned}
\end{equation}
Equations \eqnrefs{psiBzGquatxy}{psiBzGquatconstruct} are then used as before.

\subsection{Rotation Composition from Mismatched Yaw and Tilt}
\seclabel{comp_mismatch_yaw_tilt}

As an extension to the standard composition of yaw and tilt, composition is 
possible and well-defined even if the yaw and tilt are expressed relative to 
different frames. Given two frames, \fr{G} and \fr{H}, in general there is a 
unique frame \fr{B} that has a desired fused yaw $\compb{G}{\psi}$ relative to 
\fr{G}, and a desired tilt rotation component $\rotbs{H}{}{q}{t}$ relative to 
\fr{H}. Exceptions where there are multiple solutions are discussed later. 
Suppose we are given $\compb{G}{\psi}$, $\rotb{G}{H}{q} 
\mspace{-2mu}=\mspace{-2mu} (\mspace{-1mu}w_G,x_G,y_G,z_G\mspace{-1mu})$, and 
any rotation $\rotb{H}{C}{q} = (w_C,x_C,y_C,z_C)$ that has the required tilt 
rotation component $\rotbs{H}{}{q}{t}$ relative to \fr{H}. The fused yaw of 
$\rotb{H}{C}{q}$ relative to \fr{H} is irrelevant, and $\rotb{H}{C}{q}$ can in 
general be calculated directly from the tilt angles specification 
$T(0,\gamma,\alpha)$. We first calculate the cross terms
\begin{equation}
\begin{aligned}
a &= x_G x_C + y_G y_C, &\mspace{24mu} b &= x_G y_C - y_G x_C, \\
c &= w_G z_C + z_G w_C, &\mspace{24mu} d &= w_G w_C - z_G z_C.
\end{aligned}
\end{equation}
Then, using the abbreviated notation $c_{\bar{\psi}} = \cos\bigl(\tfrac{1}{2} 
\compb{G}{\psi}\bigr)$ and $s_{\bar{\psi}} = \sin\bigl(\tfrac{1}{2} 
\compb{G}{\psi}\bigr)$, we compute the following terms:
\begin{equation}
\begin{aligned}
A &= d - a, &\mspace{24mu} B &= b - c, \\
C &= b + c, &\mspace{24mu} D &= d + a, \\
G &= D c_{\bar{\psi}} - B s_{\bar{\psi}}, &\mspace{24mu} H &= A s_{\bar{\psi}} - C c_{\bar{\psi}}.
\end{aligned}
\end{equation}
The yaw rotation relative to \fr{H} from \fr{C} to \fr{B} is then given by the 
\emph{referenced rotation} (see \secref{referenced_rotations})
\begin{equation}
\rotb{HC}{B}{q} =
\begin{cases}
(\tfrac{G}{F}, 0, 0, \tfrac{H}{F}), & \text{if $F \neq 0$,} \\
(1, 0, 0, 0), & \text{otherwise,}
\end{cases}
\eqnlabel{quatrefrot}
\end{equation}
where $F = \sqrt{G^2 + H^2}$. Relative to frames \fr{H} and \fr{G}, frame \fr{B} 
is then given by the expressions
\begin{align}
\rotb{H}{B}{q} &= \rotb{HC}{B}{q} \, \rotb{H}{C}{q}, & \rotb{G}{B}{q} &= \rotb{G}{H}{q} \mspace{1mu} \rotb{HC}{B}{q} \, \rotb{H}{C}{q}. \eqnlabel{quatfromyawtilt}
\end{align}
Problems occur when $F = 0$, which occurs exactly when
\begin{equation}
\alpha_G + \alpha_C = \pi,
\end{equation}
where $\alpha_G$, $\alpha_C$ are the tilt angles of $\rotb{G}{H}{q}$, 
$\rotb{H}{C}{q}$, respectively. In this case, every \fr{B} that has the required 
tilt rotation component relative to \fr{H} has the same fused yaw relative to 
\fr{G}, namely $\yawof{\rotb{G}{C}{q}}$. As such, if $\compb{G}{\psi} = 
\yawof{\rotb{G}{C}{q}}$ then there are infinitely many solutions---of which one 
is returned by this method---but otherwise there are no solutions.

\section{Application Examples}
\seclabel{application_examples}

\begin{figure}[!t]
\parbox{\linewidth}{\centering\includegraphics[width=0.90\linewidth]{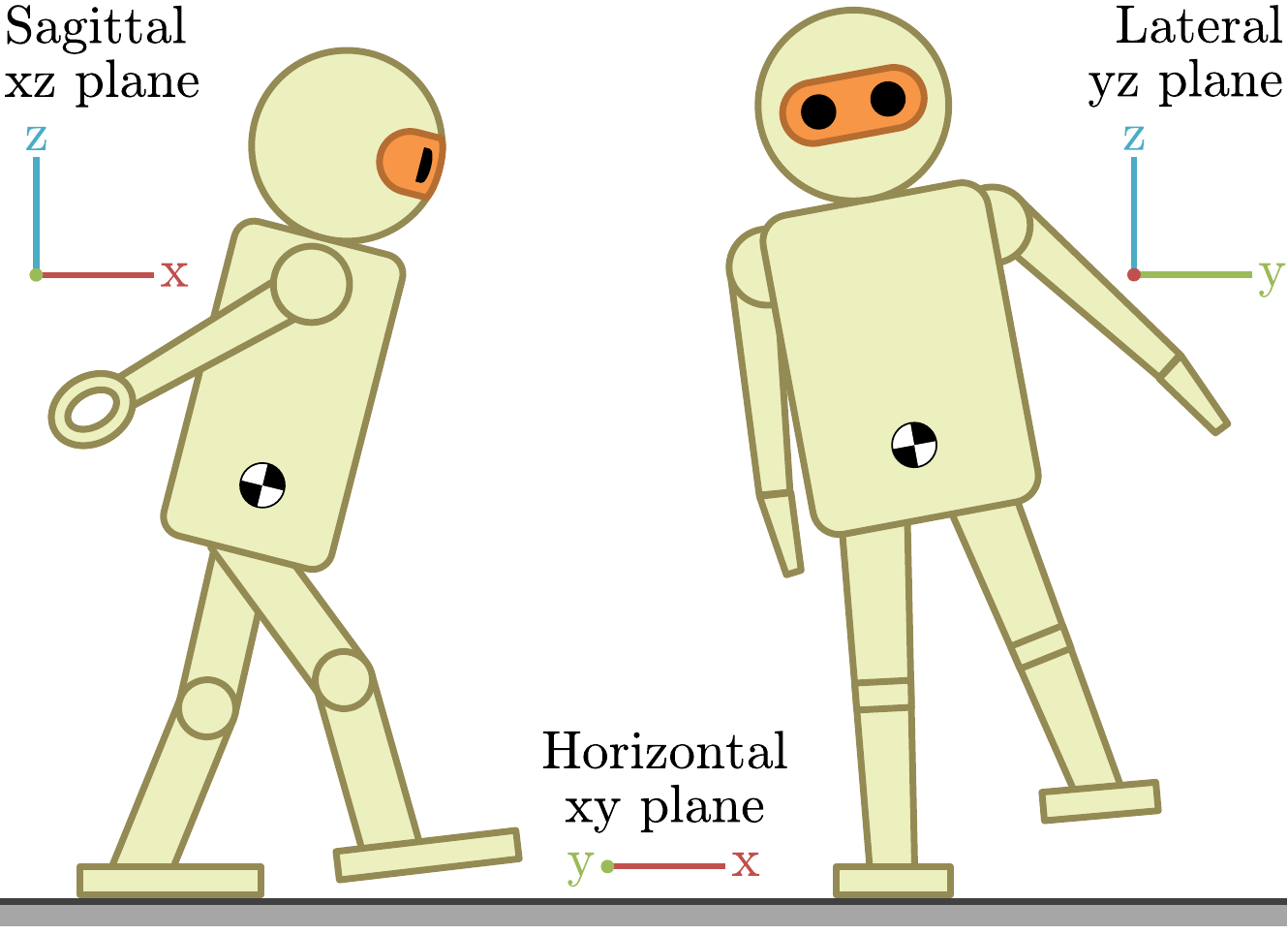}}
\caption{The relevant major planes for the application of bipedal walking. The 
fused yaw parameter $\psi \equiv p_z$ describes the heading of the robot, the 
fused pitch $\theta$ and tilt phase $p_y$ relate to the sagittal component of 
orientation of the robot, and the fused roll $\phi$ and tilt phase $p_x$ relate 
to the lateral component of orientation of the robot. Using these parameters 
allows the motion, stability and state of balance of the robot to be evaluated, 
quantified and controlled separately in the three major planes.}
\figlabel{robot_gait_application}
\vspace{-2ex}
\end{figure}

Tilt rotations and the tilt phase space can be applied, with advantages, in many 
scenarios. For example, quadrotors in general need to tilt in the direction they 
wish to accelerate, so a smooth and axisymmetric way of representing such tilt, 
using a suitable mix of tilt angles, fused angles, and in particular the tilt 
phase space, can be of great benefit. Similar arguments for the use of these 
representations also apply to the scenarios of balance and bipedal locomotion, 
where the tilt rotation component is particularly relevant because it 
encapsulates the entire heading-independent balance state of the robot, with no 
extra component of rotation about the z-axis \cite{Allgeuer2018a}. In fact, a 
tilted IMU accelerometer measures the direction of gravity, which is a direct 
measurement of $\vecbs{B}{z}{G}$, the z-vector parameterisation of tilt 
rotations. This supports the notion that tilt rotations are a natural and 
meaningful split of orientations into yaw and tilt. 
\figref{robot_gait_application} illustrates how for the application of bipedal 
walking the fused angles $\psi$, $\theta$, $\phi$ and/or the tilt phase space 
parameters $p_x$, $p_y$, $p_z$ can be used to independently quantify the amount 
of rotation in each of the three major planes. This allows the motion, stability 
and state of balance to be measured and controlled separately in each of these 
three planes, which correspond to the sagittal, lateral and horizontal, i.e.\ 
heading, planes of walking.

Due to the many advantageous properties of the tilt phase space (see 
\secref{fundamental_properties}), and in particular due to magnitude 
axisymmetry, the tilt phase space was chosen as the basis of a developed 
feedback controller for the stabilisation of bipedal walking \cite{PhaseFeed}. 
It is in the nature of feedback controllers, e.g.\ PID-style controllers, to 
produce control inputs of arbitrary magnitude based on a set of gains, so the 
unbounded nature of the tilt phase space was able to be utilised to full effect.

The concepts of tilt rotations and the tilt phase space have also been applied 
in the keypoint gait generator that underlies the feedback gait presented in 
\cite{PhaseFeed}. Most notably, the tilt phase space is used to separate the yaw 
and tilt of the feet at each gait keypoint, and then interpolate between them 
using the orientation cubic spline interpolation described in 
\secref{vector_space}. This ensures that the yaw and tilt profiles are 
individually exactly as intended in the final 3D foot trajectories, especially 
seeing as the yaw profiles come from the commanded step sizes, and the tilts 
come separately from the feedback controller. Tilt vector addition is also 
required, because multiple feedback components need to contribute to each final 
foot tilt. The summated foot tilts are not guaranteed to be in range though, so 
the unbounded nature of the tilt phase space helps in being robust to `wrapping 
around' prior to saturation. All this allows effective and robust gait 
trajectories to be generated for robots using this method.

\section{Conclusion}
\seclabel{conclusion}

Both tilt rotations and the novel tilt phase space were formally introduced in 
this paper. As was shown in detail, tilt rotations possess many remarkable 
properties, and are useful for the analysis of the rotations of rigid bodies, in 
particular balancing bodies. The value of the tilt phase space was demonstrated, 
firstly in that it provides a tool to study unbounded tilt rotations in a way 
analogous to fused angles for orientations, and secondly in that it formalises a 
vector space of tilt rotations that allows for intuitive commutative addition. 
The tilt phase space also enables more complex operations, such as for example 
cubic spline interpolation in a way that cleanly respects the independence of 
yaw and tilt. As a final note, all rotation representations that were presented 
in this paper have open source software library support, in both \cpp and Matlab 
(see \secref{introduction}).

\bibliographystyle{IEEEtran}
\bibliography{IEEEabrv,ms}

\end{allowdisplaybreaks}
\end{document}